\documentclass[11pt]{article}
\usepackage{fullpage}
\usepackage[utf8]{inputenc}
\usepackage[english]{babel}
\usepackage[T1]{fontenc}

\usepackage{amsmath}

\usepackage{amsfonts}
\usepackage{url}
\usepackage{fullpage}
\usepackage{amssymb}
\usepackage{graphicx}
\usepackage[labelfont=bf]{caption}
\def\NN{\mathrm{NN}}
\def\dt{\mathrm{d}t}

\def\Vor{\mathrm{Vor}}
\def\Pow{\mathrm{Pow}}
\def\CK{\mathrm{CK}}

 \def\calX{\mathcal{X}}
 \def\calY{\mathcal{Y}}

 \def\calF{\mathcal{F}}
\def\calS{\mathcal{S}}
\def\calB{\mathcal{B}}
\def\calD{\mathcal{D}}
\def\calY{\mathcal{Y}}

\def\bbD{\mathbb{D}}

\def\bbR{\mathbb{R}}

\def\Log{\mathrm{Log}}

\def\inner#1#2{{\langle #1,#2\rangle}}

\def\st{\ \mid\ }
\def\Bi{\mathrm{Bi}}

\def\arctanh{\mathrm{arctanh}}
\def\arccosh{\mathrm{arccosh}}

\def\cosh{\mathrm{cosh}}

\graphicspath{{./figs/}}

\def\calP{\mathcal{P}}
\def\calS{\mathcal{S}}
\def\calR{\mathcal{R}}
\def\calC{\mathcal{C}}
\def\bbR{\mathbb{R}}

\def\Bi{\mathrm{Bi}}
\def\sign{\mathrm{sign}}
\def\diag{\mathrm{diag}}

\def\st{ \ : \ }

\def\dist{\mathrm{dist}}

\def\arccosh{\mathrm{arccosh}}
\def\arctanh{\mathrm{arctanh}}

\def\push{\mathrm{push}}
\def\pull{\mathrm{pull}}

\def\angle{\mathrm{angle}}

\def\inner#1#2{\langle #1 , #2 \rangle}

\def\diag{\mathrm{diag}}

\author{Frank Nielsen\thanks{\'Ecole Polytechnique, France, and Sony Computer Science Laboratories, Japan. e-mail:{\tt Frank.Nielsen@acm.org}} \and Boris Muzellec\thanks{\'Ecole Polytechnique, France.} \and Richard Nock\thanks{Data61, The Australian National University (ANU) \& The University of Sydney, Australia.}}

\title{Large Margin Nearest Neighbor Classification  using\\ Curved Mahalanobis Distances\thanks{A preliminary work appeared at IEEE International Conference on Image Processing (ICIP) 2016~\cite{CKLMNN-2016}.}}

\date{}
\begin{document}

\maketitle

\begin{abstract}
We consider the supervised classification problem of machine learning in Cayley-Klein projective geometries:
We show how to learn a curved Mahalanobis metric distance corresponding to either the hyperbolic geometry or the elliptic geometry using the Large Margin Nearest Neighbor (LMNN) framework. We report on our experimental results, and further consider the case of learning a mixed curved Mahalanobis distance.
Besides, we show that the Cayley-Klein Voronoi diagrams are affine, and can be built from an equivalent (clipped) power diagrams, and that Cayley-Klein balls have Mahalanobis shapes with displaced centers.
\end{abstract}

\noindent {Keywords}: classification; metric learning; Cayley-Klein metrics; LMNN; Voronoi diagrams.
 
\section{Introduction}

\subsection{Metric learning}
The  {\em Mahalanobis distance} between point $p$ and $q$ of $\bbR^d$ is defined for a symmetric positive definite matrix $Q\succ 0$ by:
\begin{equation}
d_M(p,q)=\sqrt{(p-q)^\top  Q (p-q)}.
\end{equation}
It is a {\em metric} distance that satisfies the three metric axioms: indiscernibility ($d_M(p,q)=0$ iff. $p=q$), symmetry ($d_M(p,q)=d_M(q,p)$), and  triangle inequality ($d_M(p,q)+d_M(q,r) \geq d_M(p,r)$).
The Mahalanobis distance generalizes the Euclidean distance by choosing $Q=I$, the identity matrix: $D_I(p,q)=\|p-q\|$.
Given a finite point set $\calP=\{x_1,\ldots,x_n\}$, matrix $Q$ is often chosen as the {\em precision matrix} $\Sigma^{-1}$ where  $\Sigma$ is the {\em covariance matrix} of $\calP$:
\begin{eqnarray}
\Sigma&=&\frac{1}{n}\sum_i (x_i-\mu)(x_i-\mu)^\top, \text{with}\\
&\mu&=\frac{1}{n}\sum_i x_i.
\end{eqnarray}
$\mu$ is the center of mass of $\calP$ (called sample mean in Statistics).

In machine learning, given a {\em labeled point} set $\calP=\{(x_1,y_1),\ldots,(x_n,y_n)\}$ with $y_i\in\calY$ denoting the label of $x_i\in\calX$, 
the classification task consists in building a {\em classifier} $h(\cdot): \calX\mapsto \calY$ to tag newly unlabelled points $x$ as $y=h(x)$. 
The classification task is binary when $\calY=\{-1,1\}$, otherwise it is said multi-task.
A simple but powerful classifier consists in retrieving the $k$ nearest neighbor(s) $\NN_k(x)$ of an unlabeled query point $x$, and to associate to $x$ the dominant label of its neighbor(s). This rule yields the so-called {\em $k$-Nearest Neighbor classifier} (or $k$-NN for short).  
The  $k$-NN rule depends on the chosen distance between elements of $\calX$. 
When the distance is {\em parametric} like the Mahalanobis distance, one has to learn the appropriate distance parameter 
(eg., matrix $Q$ for the Mahalanobis distance). This hot topic of  machine learning bears the name {\em metric learning}.
Weinberger et al.~\cite{WeinbergerSaul2006} proposed an efficient method to learn a Mahalanobis distance: The  {\em Large Margin Nearest Neighbor} (LMNN) algorithm.
The LMNN algorithm was further extended to elliptic Cayley-Klein geometries in~\cite{Bi2015}.
In this work, we further extend the LMNN framework in hyperbolic Cayley-Klein geometries, and also consider mixed hyperbolic/elliptic Cayley-Klein distances.

\subsection{Contributions and outline}

We summarize our key contributions as follows:

\begin{itemize}
	\item We extend the LMNN to hyperbolic Cayley-Klein geometries (\S~\ref{sec:LMNN-H}),
	\item We introduce a linearly mixed Cayley-Klein  distance and investigate its experimental performance (\S~\ref{sec:mixedres}),
	\item We show that  Cayley-Klein Voronoi diagrams are affine and equivalent to power diagrams (\S~\ref{sec:CKVor}), and
	\item We prove that  Cayley-Klein balls have  Mahalanobis shapes with displaced centers (\S~\ref{sec:CKBall}).
	
\end{itemize}

The paper is organized as follows:
Section~\ref{sec:CKgeo} concisely introduces the basic notions of Cayley-Klein geometries and present formula for the elliptic/hyperbolic Cayley-Klein distances.
Those elliptic/hyperbolic Cayley-Klein distances are reinterpreted as curved Mahalanobis distances in Section~\ref{sec:curvedMAH}.
Section~\ref{sec:compgeom} studies some facts useful for computational geometry~\cite{CG-2000}: First, we show that the Cayley-Klein bisector is a (clipped) hyperplane, and that the Cayley-Klein Voronoi diagrams can be built from  equivalent (clipped) power diagrams (Section~\ref{sec:CKVor}).  Second, we notice that Cayley-Klein balls have Mahalanobis shapes with displaced centers (Section~\ref{sec:CKBall}).
Section~\ref{sec:LMNN} introduces the LMNN framework: First, we review LMNN for learning a squared Mahalanobis distance in \S~\ref{sec:LMNN-M}.
Then we report the extension of Bi et al.~\cite{Bi2015} to elliptic Cayley-Klein geometries, and describe our novel extension to hyperbolic Cayley-Klein geometries in \S~\ref{sec:LMNN-H}. Experimental results are presented in Section~\ref{sec:res}, and a  mixed Cayley-Klein distance is considered in \S~\ref{sec:mixedres} that further improve experimentally classification performance. Fast nearest neighbor queries in Cayley-Klein geometries are briefly touched upon in  \S~\ref{sec:proximity}.
Finally, Section~\ref{sec:concl} concludes this work and hints at further perspectives of the role of Cayley-Klein distances in machine learning.

\section{Cayley-Klein geometry\label{sec:CKgeo}}

The {\em real projective space}~\cite{RichterGebert2011} $\mathbb{RP}^d$ can be understood as the set of lines passing through the origin of the vector space $\mathbb{R}^{d+1}$. Projective spaces is different from spherical geometry because antipodal points of the unit sphere are identified (they yield the same line passing through the origin).
Let $\mathbb{RP}^d=(\mathbb{R}^{d+1} \backslash\{0\}) / \tilde{\ }$ denote the {real projective space} with the equivalence class relation $\sim$: $(\lambda x, \lambda) \sim (x,1)$ for $\lambda\not =0$.
A point $x$ in $\mathbb{R}^d$ is mapped to a point $\tilde{x}\in \mathbb{RP}^d$ using {\em homogeneous coordinates} $x\mapsto \tilde{x}=(x,w=1)$ by adding an extra coordinate $w$.
Conversely, a projective point $\tilde{x}\in\mathbb{R}^{d+1}=(x,w)$ is {\em dehomogeneized} by ``perspective division'' 
$\tilde{x}\mapsto \frac{x}{w}\in\mathbb{R}^{d}$ provided that $w\not=0$. The projective points at infinity have the coordinate $w=0$. Thus the projective space is a compactification of the Euclidean space.
The non-infinite points  of the projective space $\mathbb{RP}^d$ is often visualized in $\mathbb{R}^{d+1}$ as the points lying on the hyperplane $H$ passing through the $(d+1)$-th coordinate $w=1$ (with each point on $H$ defining a line passing through the origin of $\mathbb{R}^{d+1}$).
In projective geometry, two distinct lines always intersect in exactly one point, and a bundle of Euclidean parallel lines intersect at the same projective point at infinity.

\begin{figure}
\begin{center}
\begin{tabular}{cc}
\includegraphics[width=0.4\textwidth]{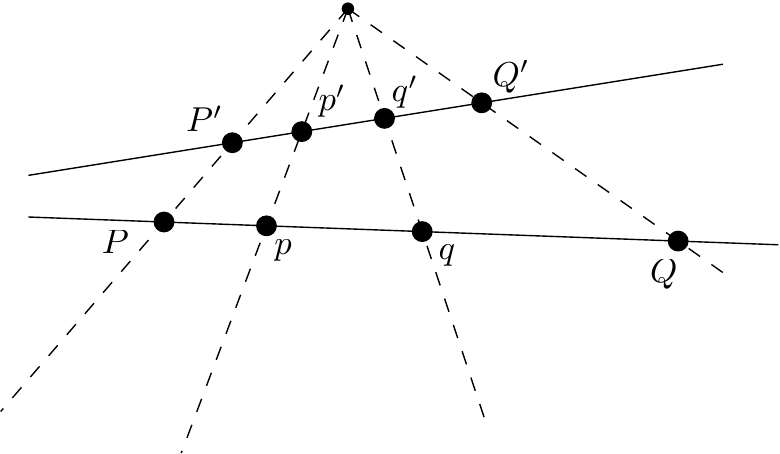} &
\includegraphics[width=0.4\textwidth]{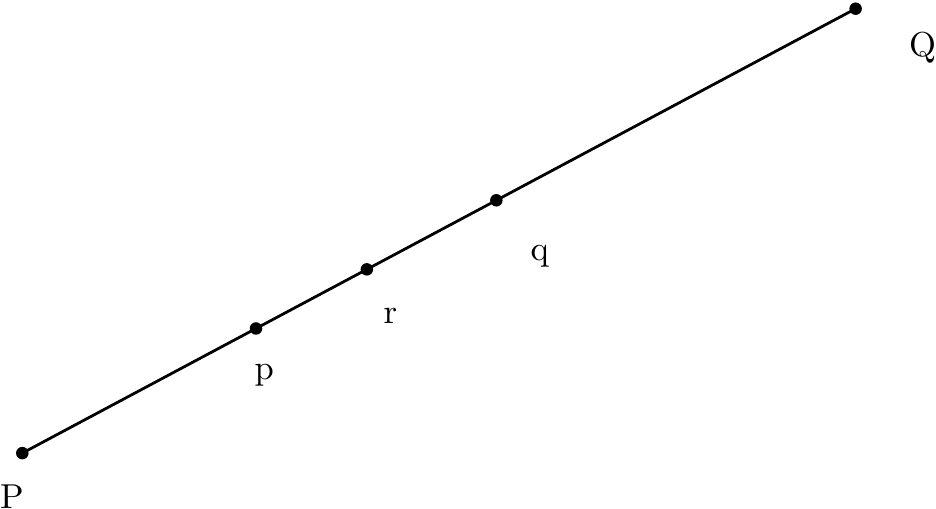}\\
(a) & (b)
\end{tabular}
\end{center}
\caption{Cross-ratio: (a) The cross-ratio $(p,q;P,Q)$ of four collinear points $p,q,P,Q$ is invariant under a collineation: $(p,q;P,Q)=(p',q';P',Q')$.
 (b) The cross-ratio satisfies the identity $(p,q;P,Q) = (p,r;P,Q) \times (r,q;P,Q)$ when $r$ is collinear with $p,q,P,Q$.
\label{fig:cr}}
\end{figure}

In projective geometry~\cite{RichterGebert2011}, the {\em cross-ratio} (Figure~\ref{fig:cr})  of {\em four collinear points} $p,q,P,Q$ on a line is defined by:
	
\begin{equation}
(p,q;P,Q) = \frac{(p-P)(q- Q)}{(p-Q)(q- P)}.
\end{equation}

The cross-ratio is a {\em measure} that is {\em invariant} by projectivities~\cite{RichterGebert2011} (see Figure~\ref{fig:cr} (a)), also called collineations or homographies.
The cross-ratio enjoys the following key properties:

\begin{itemize}
\item $(p,p;P,P) = 1$,  
\item $(p,q;Q,P) = \frac{1}{(p,q;P,Q)}$,  
\item $(p,q;P,Q) = (p,r;P,Q) \times (r,q;P,Q)$ when $r$ is collinear with $p,q,P,Q$.
\end{itemize}

A gentle introduction to projective geometry and Cayley-Klein geometries can be found in~\cite{RichterGebert2011,struve-2004,struve-2010}.
We also refer the reader to a more advanced textbook~\cite{CK-2006} handling invariance and isometries, and to the historical seminal paper~\cite{Cayley-1859} of Cayley (1859).

\subsection{Cayley-Klein distances from cross-ratio measures}

A {\em Cayley-Klein geometry} is a {\em triple} $\mathcal{K} = (\mathcal{F},c_{\dist}, c_{\angle})$, where:
 
	\begin{enumerate}
	\item  $\mathcal{F}$  is a {\em fundamental conic},

	\item $c_{\dist} \in \mathbb{C}$ is a constant unit for measuring distances, and

	\item $c_{\angle} \in \mathbb{C}$ is constant unit  for measuring angles. 
	\end{enumerate}

\begin{figure}
\begin{center}
\begin{tabular}{cc}
\includegraphics[height=4cm]{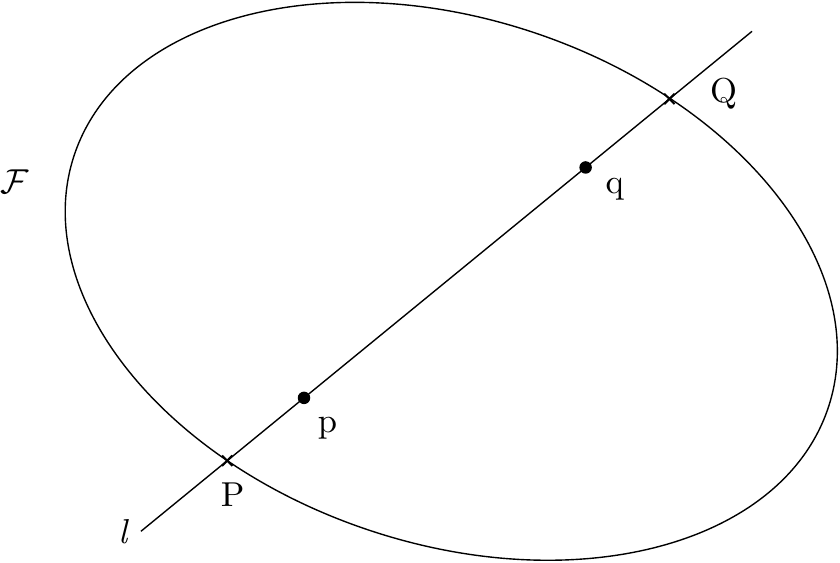} & \includegraphics[height=4cm]{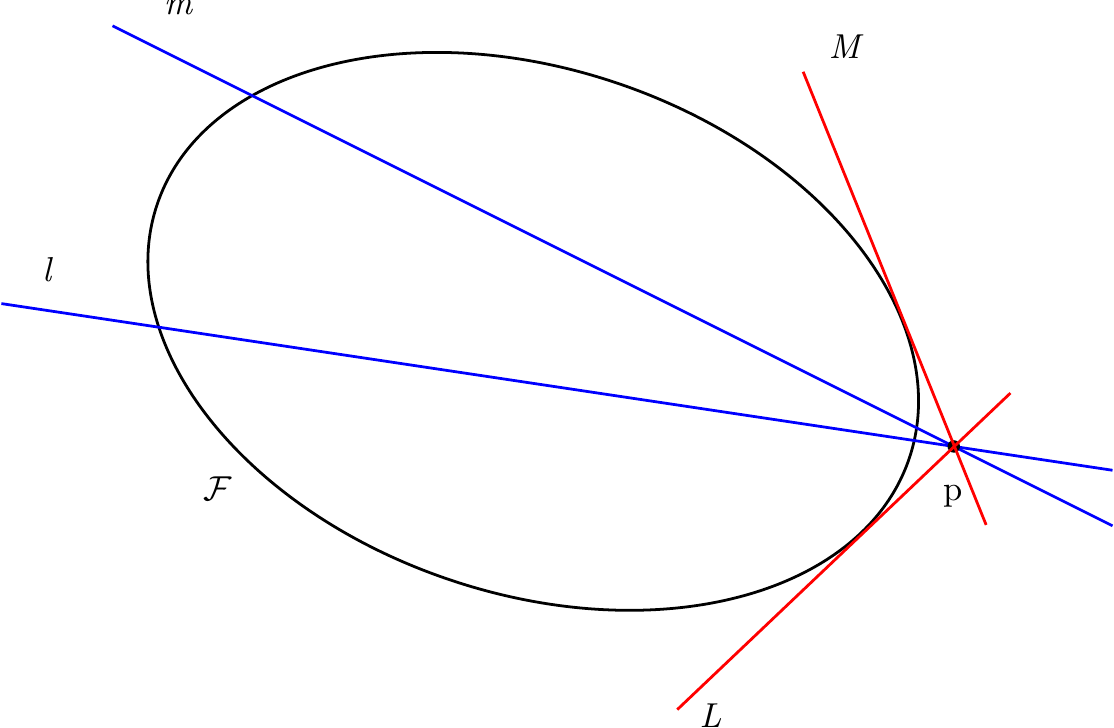}\\
(a) distance measurement & (b) angle measurement\\
 $\dist(p,q) = c_{\dist}\ \Log((p,q;P,Q))$ & $\angle(l,m) = c_{\angle}\ \Log((l,m;L,M))$
\end{tabular}
\end{center}
\caption{Distance and angle measurements in Cayley-Klein geometry.~\label{fig:distangle}}
  \end{figure}

The distance in Cayley-Klein geometries (see Figure~\ref{fig:distangle}) is defined by:
\begin{equation}
\dist(p,q) = c_{\dist}\ \Log((p,q;P,Q)), \label{eq:Dlog}
\end{equation}
where $P$ and $Q$ are the intersection points of line $l=(pq)$  with the  fundamental conic $\calF$.
Historically, the fundamental conic was called the ``absolute''~\cite{Cayley-1859}.
The logarithm function $\Log$ denotes the {\em principal value} of the {\em complex logarithm}.
That is, since complex logarithm values are defined up to modulo $2\pi i$, we define the principal value of the complex logarithm
as the unique value with imaginary part lying in the range $(-\pi,\pi]$.

Similarly, the angle in Cayley-Klein geometries (see Figure~\ref{fig:distangle}) is measured as follows:
\begin{equation}
\angle(l,m) = c_{\angle}\ \Log((l,m;L,M)), \label{eq:Alog}
\end{equation}
where $L$ and $M$ are tangent lines to the fundamental conic $\calF$ passing through the intersection point $p$ of  line $l$ and line $m$
 (see Figure~\ref{fig:distangle}).
This formula generalizes the Laguerre formula that calculates the acute angle between two distinct real lines~\cite{RichterGebert2011}.

The Cayley-Klein geometries can further be extended to Hilbert projective geometries~\cite{Hilbert-2004} by replacing the conic object $\calF$ with a bounded convex subset of $\bbR^d$. Interestingly, the convex objects delimiting the Hilbert geometry domain do not need to be strictly convex~\cite{HilbertPolygonal-2011}.

The properties of Cayley-Klein distances are:
\begin{itemize}
\item  Law of the indiscernibles:  $\dist(p,q) = 0$ iff. $p=q$,
\item  Signed distances : $\dist(p,q) = -\dist(q,p)$, and
\item  When $p,q,r$ are collinear, $\dist(p,q) = \dist(p,r) + \dist(r,q)$.
That is, shortest-path geodesics\footnote{Cayley-Klein geometries can also be studied from the viewpoint of Riemannian geometry.} in Cayley-Klein geometries are  {\em straight lines} (clipped within the conic domain $\mathbb{D}$).
\end{itemize}	

Notice that the logarithm of Cayley-Klein measurement formula is transferring {\em multiplicative properties} of the cross-ratio to {\em additive properties} of Cayley-Klein distances:
For example, it follows from the cross-ratio identity  $(p,q;P,Q) = (p,r;P,Q) \times (r,q;P,Q)$ for collinear  $p,q,P,Q$   that
$\dist(p,q) = \dist(p,r) + \dist(r,q)$.

\subsection{Dual conics and taxonomy of Cayley-Klein geometries}

In projective geometry, {\em points} and {\em lines} are dual concepts, and theorems on points can be translated equivalently to theorems on lines.
For example, Pascal's theorem is dual to Brianchon's theorem~\cite{RichterGebert2011}.

A conic object $\calF$ can be described as the convex hull of its extreme points (points lying on its border), or equivalently as the intersection of all half-spaces tangent at its border and fully containing the conic. This is similar to the dual  $H$-representation and $V$-representation of finite convex polytopes~\cite{cvxpolytope-2013} ('H' standing for Halfspaces, and 'V' for Vertex).
This point/line duality yields a dual parameterizations of the fundamental conic $\mathcal{F} = (A,A^{\Delta})$ by two matrices, 
where  $A^\Delta=A^{-1}|A|$ is the {\em adjoint matrix} (transpose of its cofactor matrix).
Observe that the adjoint matrix can be computed even when $A$ is not invertible ($|A|=0$).

To a symmetric positive semi-definite matrix $(d+1)\times (d+1)$-dimensional $A$, we associate a {\em homogeneous polynomial} called the {\em quadratic form} $Q_A(x)=\tilde{x}^\top A \tilde{x}$.
The primal conic is thus described as the set of border points  $\calC_A=\{ \tilde{p}\in\mathbb{RP}^d \st Q_A(\tilde{p})=0\}$ using matrix $A$, and
the dual conic as the set of tangent hyperplanes $\calC_A^*=\{ \tilde{l}\in\mathbb{RP}^d  \st Q_{A^\Delta}(\tilde{l})=0\}$ using the dual adjoint matrix $A^\Delta$.

The {\em signature of matrix} is a triple $(n,z,p)$ counting the signs of the eigenvalues (in $\{-1,0,+1\}$) of its eigendecomposition, 
where $n$ denotes the number of negative eigenvalue(s), $0$ the number of null eigenvalue(s), and $p$ the number of positive eigenvalue(s) (with $n+z+p=d+1$).
For example, a $(d+1)\times (d+1)$ symmetric positive-definite matrix $S\succ 0$ has signature $(0,0,d+1)$, 
while a semi-definite rank-deficient matrix $S\succeq 0$ of rank $r<d+1$ has signature $(0,d+1-r,r)$.

Table~\ref{tab:2Dtypo} displays the seven types of planar Cayley-Klein geometries (induced by a pair of $3\times 3$ dual conic matrices $(A,A^{\Delta})$).
All degenerate cases can be obtained as the limit of non-degenerate cases, see~\cite{RichterGebert2011}. 
Another way to classify the Cayley-Klein geometries is to consider the type of measurements for distances and angles.
Each type of measurement is of three kinds~\cite{RichterGebert2011}: elliptic or hyperbolic for non-degenerate geometries or {\em parabolic} for degenerate cases.
Using this classification, we obtain nine  combinations for the planar Cayley-Klein geometries.

\begin{table}[ht]
\renewcommand{\arraystretch}{1.5}
\begin{center}

  \resizebox{\linewidth}{!}{
  \begin{tabular}{|c|c|c|c|}
  \hline
  Type &$A$ &$A^{\Delta}$ & Conic in $\mathbb{RP}^2$\\
  \hline
  \textbf{Elliptic} & $(+,+,+)$ & $(+,+,+)$ & non-degenerate complex conic  \\
  \hline
  \textbf{Hyperbolic} & $(+,+,-)$ & $(+,+,-)$ & non-degenerate real conic \\
  \hline
 Dual Euclidean & $(+,+,0)$ & $(+,+,0)$ & Two complex lines with a real intersection point \\
  \hline
  Dual Pseudo-euclidean & $(+,-,0)$ & $(+,0,0)$ & Two real lines with a double real intersection point\\
  \hline
  \textbf{Euclidean} & $(+,0,0)$ & $(+,+,0)$ & Two complex points with a double real line passing through\\
  \hline
  Pseudo-euclidean & $(+,0,0)$ & $(+,-,0)$ & Two complex points with a double real line passing through\\
  \hline
  Galilean & $(+,0,0)$ & $(+,0,0)$ & Double real line with a real intersection point\\
  \hline
  \end{tabular}}
  \end{center}
	\caption{Taxonomy of the seven planar Cayley-Klein geometries.  \label{tab:2Dtypo}}
\end{table}

Traditionally, hyperbolic geometry~\cite{HG-2006}  considers objects \textit{inside} the unit ball in the Beltrami-Klein model.
In that case, the fundamental conic that is the unit ball.
However, using Cayley-Klein geometry, {\em complex-valued measures} are also possible even when points/lines fall outside the fundamental conic. 
With the following choice $c_{\dist} = -\frac{1}{2}$ and $c_{\angle} = \frac{i}{2}$, we obtain~\cite{RichterGebert2011} (Chapter 20):

\begin{itemize}
\item A real measurement for angles when points $p,q$ lie inside the primal conic,
\item When both points $p$ and $q$ lie outside the conic, with $l=(pq)$ denoting the line passing through them:
	\begin{itemize}
	\item A real hyperbolic measure  if $l$ does not intersect the conic,
	\item A pure imaginary elliptical measure  if $l$ does not intersect the conic,
	\end{itemize}
\item A complex  measure  ($a +ib$) if one point is inside, and the other outside the conic.
\end{itemize}

Therefore it may be convenient to use in general the module of Cayley-Klein measures to handle all those possible situations.

In higher dimensions~\cite{Gunn2011,RichterGebert2011}, Cayley-Klein geometries unify common {\em space} geometries (euclidean, elliptical, and hyperbolic) with other {\em space-time} geometries (Minkowskian, Galilean, de Sitter, etc.)
In the remainder, we consider the non-degenerate {\em hyperbolic  Cayley-Klein geometry} (signature $(0,0,d+1)$, a real conic) and the non-degenerate {\em elliptic Cayley-Klein geometry} (signature $(1,0,d)$, a complex conic).

\subsection{Bilinear form and formula for the hyperbolic/elliptic Cayley-Klein distances}

For getting real-value Cayley-Klein distances, we choose the  constants as follows  (with $\kappa$ denoting the  curvature) :
	\begin{itemize}
	\item  Elliptic  ($\kappa>0$): $c_{\dist} = \frac{\kappa}{2i}$, 
	\item  Hyperbolic ($\kappa<0$): $c_{\dist} = -\frac{\kappa}{2}$.
  \end{itemize}  
	
By introducing the {\em bilinear form} for a $(d+1)\times (d+1)$ matrix S:
\begin{equation}
S_{pq} = (p^\top,1)^\top S (q,1) = \tilde{p}^\top S \tilde{q},
\end{equation} 
we    get rid of the \textit{cross-ratio} expression in distance/angle formula of Eq.~\ref{eq:Dlog} and Eq.~\ref{eq:Alog} using~\cite{RichterGebert2011}: 

\begin{equation}
(p,q;P,Q) = \frac{S_{pq}+\sqrt{S_{pq}^2 - S_{pp}S_{qq}}}{S_{pq}-\sqrt{S_{pq}^2 - S_{pp}S_{qq}}}.
\end{equation}

Thus, we end up with the following equivalent expressions for the elliptic/hyperbolic Cayley-Klein distances:
\begin{description}

\item[Hyperbolic Cayley-Klein distance.]
When $p,q \in \mathbb{D}_S= \lbrace p : S_{pp} <0 \rbrace$ (the hyperbolic domain), we have the following equivalent hyperbolic Cayley-Klein distances:

\begin{eqnarray}
d_{H}(p,q) &=& -\frac{\kappa}{2}  \log\left(\frac{S_{pq} + \sqrt{S_{pq}^2 - S_{pp} S_{qq}}}{S_{pq} - \sqrt{S_{pq}^2 - S_{pp} S_{qq}}}\right),\\
d_{H}(p,q) &=& -\kappa \ \arctanh \left(\sqrt{1 - \frac{S_{pp} S_{qq}}{S_{pq}^2}}\right),\\
d_{H}(p,q) &=& -\kappa\ \arccosh \left(\frac{S_{pq}}{\sqrt{S_{pp} S_{qq}}}\right),
\end{eqnarray}
where $\arccosh(x)=\log(x+\sqrt{x^2-1})$ and $\arctanh(x)=\frac{1}{2}\log \frac{1+x}{1-x}$.

\item[Elliptic Cayley-Klein distance.]
When $p,q \in \mathbb{R}^{d+1}$, we have   the following equivalent elliptic Cayley-Klein distances:

\begin{eqnarray}
d_{E}(p,q) &=& \frac{\kappa}{2i}\ \Log\left(\frac{S_{pq} + \sqrt{S_{pq}^2 - S_{pp} S_{qq}}}{S_{pq} - \sqrt{S_{pq}^2 - S_{pp} S_{qq}}}\right),\\
d_{E}(p,q) &=& \kappa\  \arccos \left(\frac{S_{pq}}{\sqrt{S_{pp} S_{qq}}}\right).\label{eq:DEarccos} 
\end{eqnarray}

Notice that $d_{E}(p,q) < \kappa\pi$, and that $p$ and $q$ always belong to the domain  $\mathbb{D}_S=\mathbb{R}^d$ in the case of elliptic geometry.
The link between the principal logarithm of Eq.~\ref{eq:Dlog} and the $\arccos$ function of Eq.~\ref{eq:DEarccos} is explained by
the following identity:  $\Log(x)=2i\arccos\left(\frac{x+1}{2\sqrt{x}}\right)$.
 \end{description}

Since the elliptic/hyperbolic case is induced by the signature of matrix $S$, we shall denote generically by $d_S$ the Cayley-Klein distance in either the elliptic or hyperbolic case.

\begin{figure}
\begin{center}
\begin{tabular}{cc}
\includegraphics[scale=0.8]{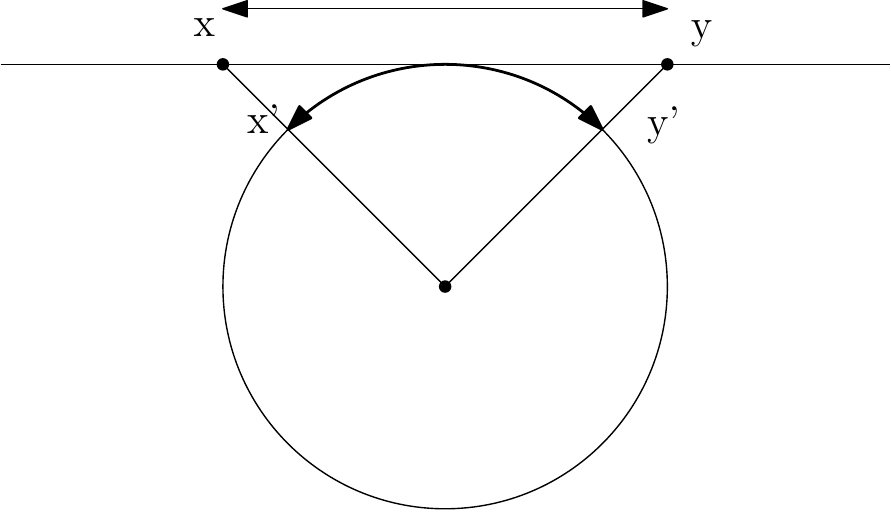} & \includegraphics[scale = 0.8]{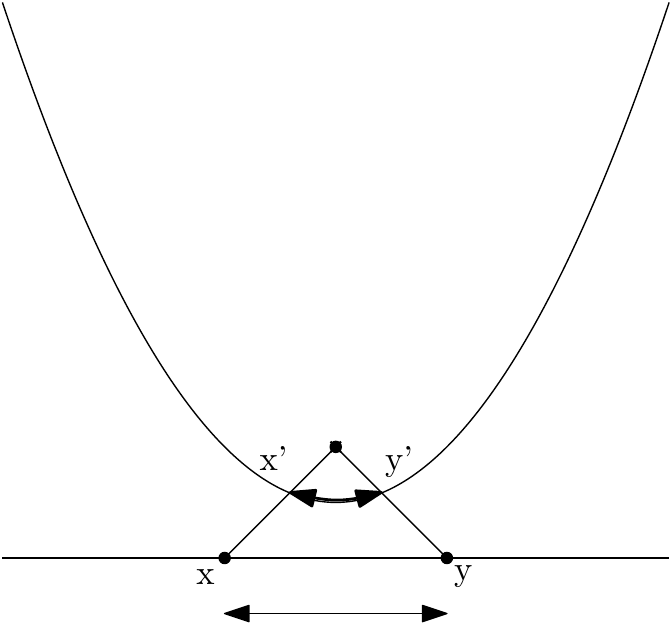}\\
(a) gnomonic projection   & (b) central projection\\
 $d_{E}(x,y) = \kappa \ \arccos \left(\inner{x'}{y'}\right)$  & $d_{H}(x,y) =\kappa\ \arccosh \left(\prec x',y'\succ\right)$\\
Euclidean inner product & Minkowski $\mathbb{R}^{d,1}$  inner product\\
$\inner{x'}{y'}=\sum_{i=1}^{d+1} x'_iy'_i$ & 
$\prec x',y'\succ=-x'_{d+1}y'_{d+1}+\sum_{i=1}^d x'_iy'_i$ \\
hemisphere model & hyperboloid model
\end{tabular}
\end{center}
\caption{Interpreting Cayley-Klein distances using projections.~\label{fig:distproj}}
  \end{figure}

Those elliptic/hyperbolic distances can be interpreted from {\em projections}~\cite{RichterGebert2011,furtherHVD-2014}, as depicted in Figure~\ref{fig:distproj}.

It is somehow surprising that we can derive metric structures from projective geometry.
Arthur Cayley (1821-1895), a British mathematician,  said ``Projective geometry is all geometry''.

\subsection{Cayley-Klein elliptic/hyperbolic distances: Curved Malahanobis distances\label{sec:curvedMAH}}

Bi et. al~\cite{Bi2015} rewrote the bilinear form as follows: 
Let 
\begin{equation}
S=\left[\begin{array}{cc}\Sigma & a \cr a^\top & b\end{array}\right]=S_{\Sigma,a,b},
\end{equation} 
with $\Sigma\succ 0$ a $d\times d$-dimensional matrix and $a,b\in\mathbb{R}^d$ so that:
\begin{equation}
S_{p,q}=\tilde{p}^\top S \tilde{q} = p^\top \Sigma q +  p^\top a + a^\top q + b.
\end{equation}

Let $\mu=-\Sigma^{-1}a\in\bbR^d$ (so that $a=-\Sigma\mu$)   and $b=\mu^\top\Sigma\mu + \sign(\kappa)\frac{1}{\kappa^2}$ so that:
\begin{equation}
\kappa=\left\{\begin{array}{ll}
(b-\mu^\top \mu)^{-\frac{1}{2}} & b> \mu^\top \mu\\
-(\mu^\top \mu-b)^{-\frac{1}{2}}  & b< \mu^\top \mu
\end{array}  \right.
\end{equation}

Then the bilinear form can be rewritten as:
\begin{equation}
S(p,q)= S_{\Sigma,\mu,\kappa}(p,q)= (p-\mu)^\top \Sigma (q-\mu) + \sign(\kappa)\frac{1}{\kappa^2}.
\end{equation}

Furthermore, it is proved in~\cite{Bi2015} that:

\begin{equation}
\lim_{\kappa\rightarrow 0^+} D_{\Sigma,\mu,\kappa}(p,q) = \lim_{\kappa\rightarrow 0^-} 
D_{\Sigma,\mu,\kappa}(p,q)  = D_{\Sigma}(p,q)
\end{equation}

Therefore the hyperbolic/elliptic Cayley-Klein distances can be interpreted as {\em curved Mahalanobis distances}  (or $\kappa$-Mahalanobis distances).
Indeed, we choose to term those hyperbolic/elliptic Cayley-Klein distances  ``curved Mahalanobis distances'' to constrast with the fact that (squared) Mahalanobis distances are symmetric Bregman divergences that induce a (self-dual) {\em flat geometry} in information geometry~\cite{IG-2016}.

Notice that when $S=\diag(1, 1, ..., 1, -1)$, we recover the canonical hyperbolic  distance~\cite{HVDeasy-2010} in Cayley-Klein model: 
\begin{equation}
D_h(p,q)=\arccosh \left(\frac{1-\inner{p}{q}}{\sqrt{1-\inner{p}{p}}\sqrt{1-\inner{q}{q}}}\right),
\end{equation}
defined inside the interior of a unit ball since we have:
\begin{equation}
S_{pq}=\begin{pmatrix}p\cr 1\end{pmatrix}^\top  \begin{pmatrix} I & 0 \cr 0 & -1  \end{pmatrix} 
\begin{pmatrix}q\cr 1\end{pmatrix}
=
p^\top I q -1 =p^\top q -1.
\end{equation}

\section{Computational geometry in Cayley-Klein geometries\label{sec:compgeom}}

\subsection{Cayley-Klein Voronoi diagrams\label{sec:CKVor}}
Define the {\em bisector} $\Bi(p,q)$ of points $p$ and $q$ as: 
\begin{equation}
\Bi(p,q) = \{x \in\mathbb{D}_S \st \dist_S(p,x) = \dist_S(x,q)\}.
\end{equation}

Then it comes that the bisector is  a hyperplane (eventually clipped to the domain $\bbD$) with equation:
\begin{eqnarray}
\lefteqn{\left\langle x,\sqrt{|S(p,p)|}\Sigma{q}  - \sqrt{|S(q,q)|}\Sigma{p} \right\rangle}\nonumber\\
&& + \sqrt{|S(p,p)|}({a}^\top({q}+x) + b)
- \sqrt{|S(q,q)|}({a}^\top({p}+x) + b) = 0
\end{eqnarray}

\begin{figure}
\begin{center}
\includegraphics[width=0.3\textwidth]{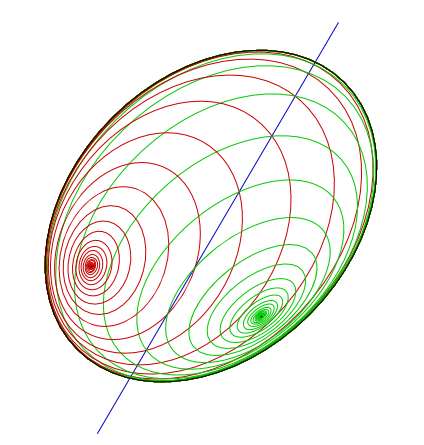}\hskip 2cm\includegraphics[width=0.3\textwidth]{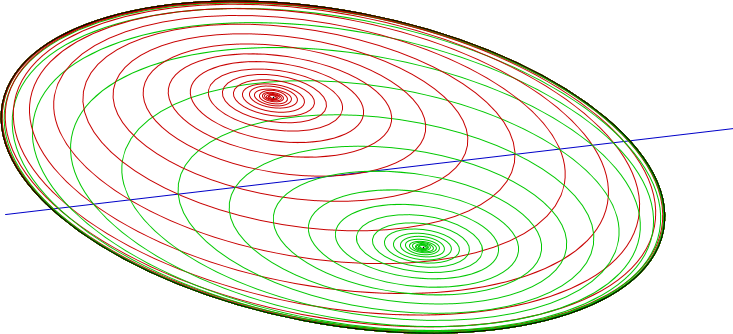}
\end{center}

\caption{Two examples of bisectors of two points in hyperbolic Cayley-Klein geometries (with the respective fundamental conic displayed in thick black).\label{fig:bi}}
\end{figure}

Figure~\ref{fig:bi} displays two examples of the bisectors of two points in planar hyperbolic Cayley-Klein geometry.

Thus the Cayley-Klein Voronoi diagram is an {\em affine diagram}.
Therefore the Cayley-Klein Voronoi diagram can be computed as an equivalent (clipped) {\em power diagram}~\cite{BVD-2007,BVD-2010,HVDeasy-2010}, using the following conversion formula:
\begin{eqnarray}
c_i &=& \frac{\Sigma p_i+ a}{2\sqrt{S_{p_ip_i}}},\\
r_i^2 &=& \frac{\Vert\Sigma p_i+a\Vert^2}{4S_{p_ip_i}} + \frac{a^\top p_i + b}{\sqrt{S_{p_ip_i}}},
\end{eqnarray}
where $B_i=(c_i,r_i)$ is the equivalent ball of point $p_i\in\calP$.

More precisely, let $\calB=\{B_i=(c_i,r_i) \st i\in [n] \}$ denote the set of associated balls of $\calP$.
Then the Cayley-Klein Voronoi diagram $\Vor_{S}^\CK(\calP)$ of $\calP$ amounts to the intersection of the power Voronoi diagram $ \Vor^\Pow(\calB)$ of equivalent balls clipped to the domain $\bbD$:

\begin{equation}
\Vor_{S}^\CK(\calP) =  \Vor^\Pow(\calB)\cap \mathbb{D}_S.
\end{equation}

\begin{figure}%
\centering
\includegraphics[width = 0.40\textwidth]{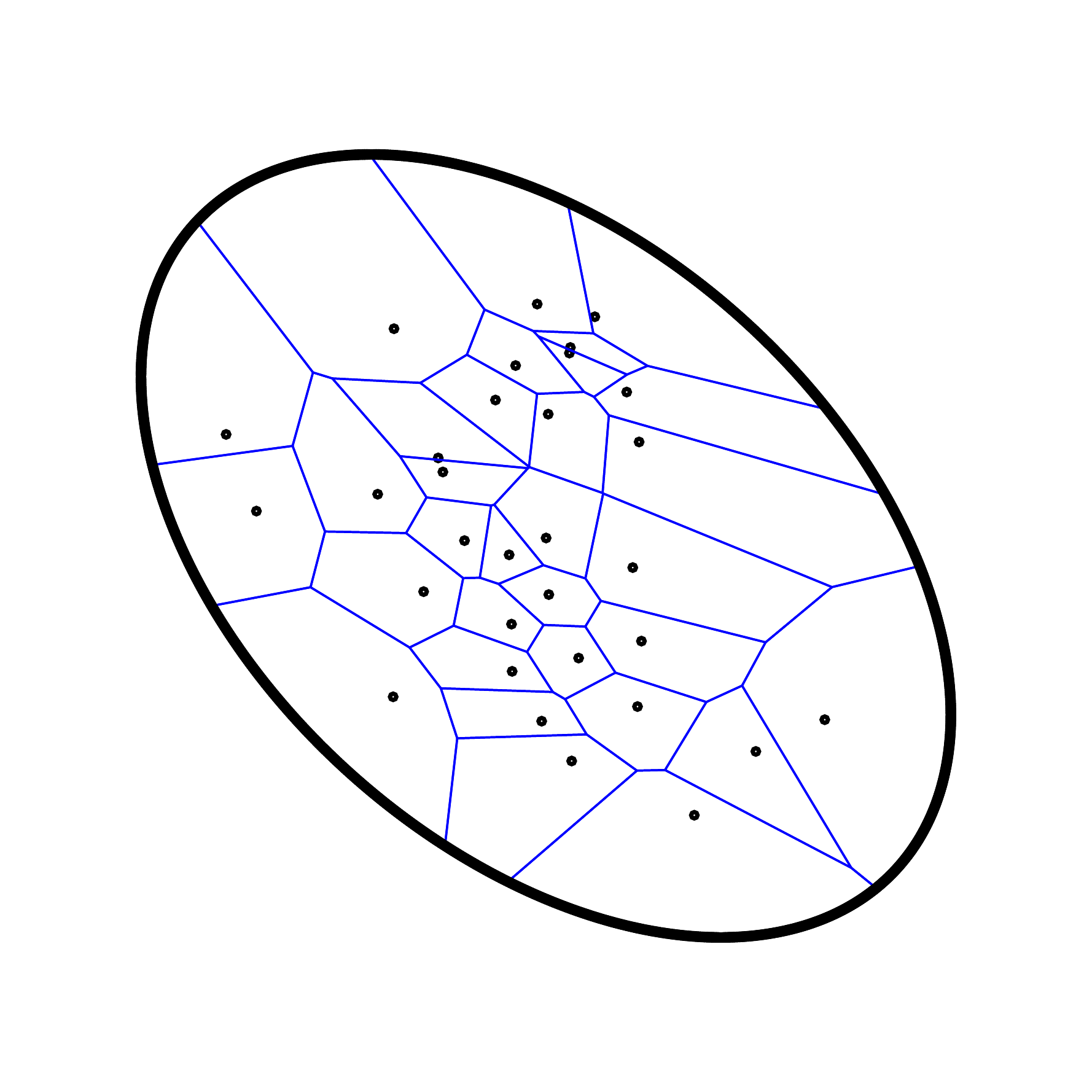} \hskip 1cm \includegraphics[width = 0.40\textwidth]{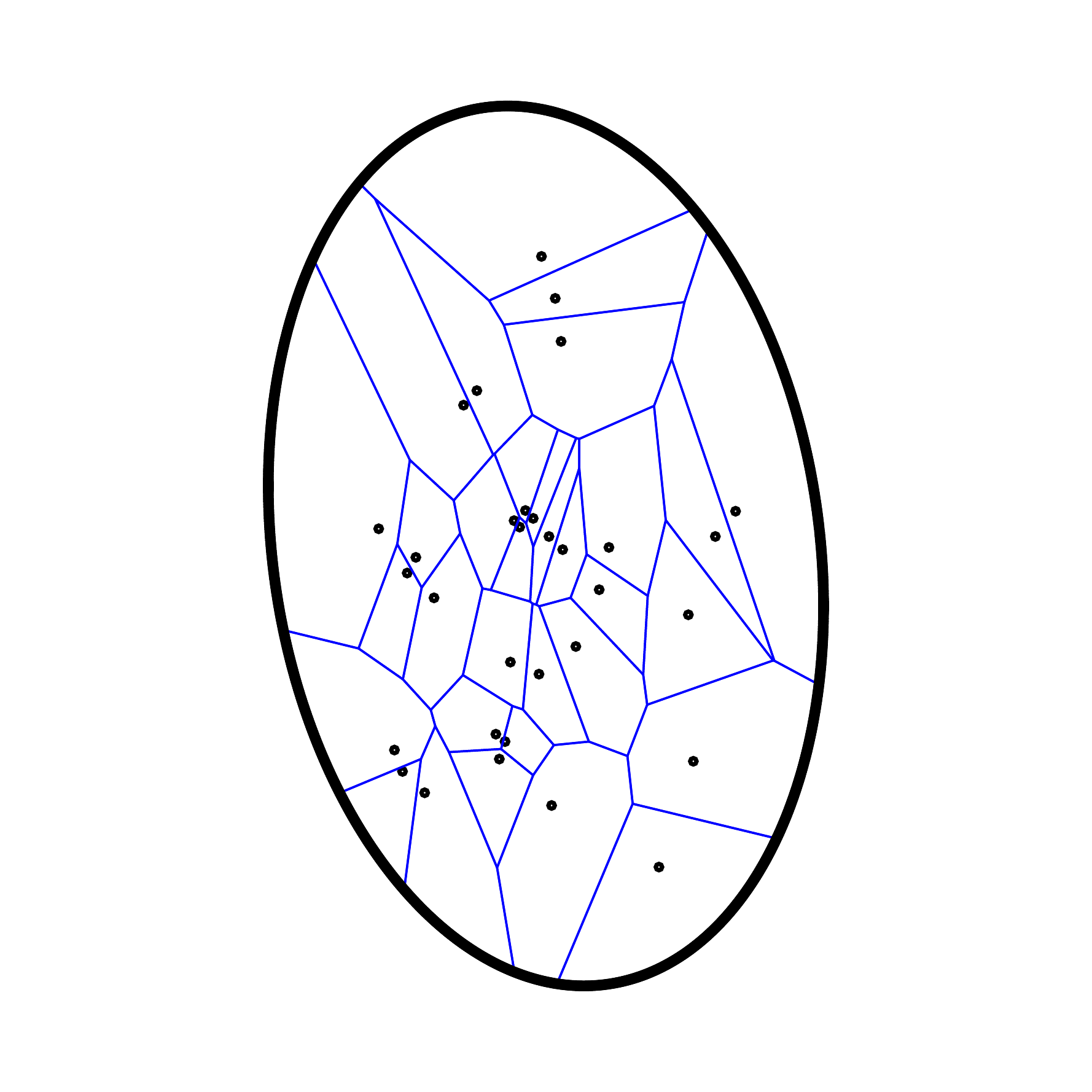} 
\caption{Example of hyperbolic Cayley-Klein Voronoi diagrams that are clipped affine diagrams.}%
\label{fig:ckvor}%
\end{figure}

Figure~\ref{fig:ckvor} and a short online video\footnote{\url{https://www.youtube.com/watch?v=YHJLq3-RL58}} illustrates the Cayley-Klein Voronoi diagrams.

\subsection{Cayley-Klein balls have Mahalanobis shapes with displaced centers\label{sec:CKBall}}

A {\em Cayley-Klein ball} $B$ of center $c$ and radius $r$ is defined by:
\begin{equation}
B^\CK(c,r)=\{x \st d_\CK(x,c)\leq r\}.
\end{equation}
The {\em Cayley-Klein sphere} $S=\partial B^\CK$ has equation $d_\CK(x,c)=r$.

\begin{figure}%
\centering
\includegraphics[scale = 0.40]{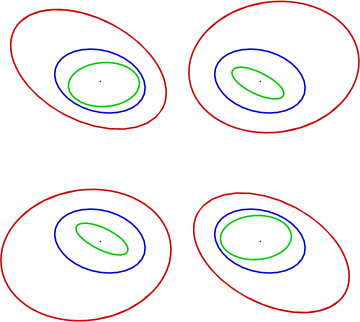} 
\caption{Cayley-Klein spheres: Elliptic (red), hyperbolic (green), and Mahalanobis spheres (blue). The dots indicates the centers of those spheres.}%
\label{fig:ckball}%
\end{figure}

Figure~\ref{fig:ckball} shows Cayley-Klein spheres in the elliptic case (red), and in the hyperbolic case (green) at different center positions (but for fixed elliptic and hyperbolic geometries).
For comparison, the Mahalanobis spheres are displayed (blue):
 This drawing let us visualize the {\em anisotropy} of  Cayley-Klein spheres that have shape depending on the center location, while Mahalanobis spheres have identical shapes everywhere ({\em isotropy}).

It can be noticed in Figure~\ref{fig:ckball} that Cayley-Klein balls have {\em Mahalanobis ball shapes} with {\em displaced centers}.
We shall give the corresponding conversion formula.
Let 
\begin{equation}
(x-c')^\top \Sigma' (x-c')={r'}^2, 
\end{equation}
denote the equation of a Mahalanobis sphere of center $c'$, radius $r'$, and shape $\Sigma'\succ 0$.
Then a hyperbolic/elliptic sphere can be interpreted as a Mahalanobis sphere as follows:

\begin{description}
\item[Hyperbolic Cayley-Klein sphere case:]
 
 \begin{equation*}
\begin{split}
\Sigma' &= a a^\top - \tilde{r}^2\Sigma\\
c' &= \Sigma'^{-1} ( \tilde{r}^2 a - b'a')\\
r'^2 &= \tilde{r}^2 b -b'^2  + \inner{c'}{c'}_{\Sigma'}\\
\end{split}
\quad \text{with} \quad
\begin{split}
\tilde{r} &= \sqrt{S_{c,c}}\cosh(r)\\
a' &= \Sigma c + a\\
b' &= a^\top c + b\\
\end{split}
\end{equation*}

\item[Elliptic Cayley-Klein sphere case:]
\begin{equation*}
\begin{split}
\Sigma' &= \tilde{r}^2\Sigma - a a^\top\\
c' &= \Sigma'^{-1} (b'a' - \tilde{r}^2 a)\\
r'^2 &= b'^2 - \tilde{r}^2 b  + \inner{c'}{c'}_{\Sigma'}\\
\end{split}
\quad  \text{with} \quad
\begin{split}
\tilde{r} &= \sqrt{S_{c,c}}\cos(r)\\
a' &= \Sigma c + a\\
b' &= a^\top c + b\\
\end{split}
\end{equation*}

\end{description}

Furthermore, by using the Cholesky decomposition of $\Sigma=L L^\top=\Sigma^\top= L^\top L$, a Mahalanobis sphere can be interpreted as an {\em ordinary Euclidean sphere} after
 performing an affine transformation $x_L\leftarrow L x$.

\begin{eqnarray}
(x-c')^\top \Sigma' (x-c') &=& {r'}^2,\\
(L(x-c'))^\top (L(x-c'))  &=&{r'}^2,\\
 (x_L-c'_L)^\top  (x_L-c'_L) &=&{r'}^2, \\
\|x_L-c'_L\|_2 &=& r'.
\end{eqnarray}

\section{Learning curved Mahalanobis metrics\label{sec:LMNN}}

Supervised learning techniques rely on labelled information, or at least on side information based on  similarities/dissimilarities. 
In  the technique called {\em Mahalanobis Metric for Clustering} (MMC)~\cite{Xing2003}, Xing and al. use {\em pairwise information} to learn a global Mahalanobis metric. 
Given two sets $\calS$ and $\calD$ of input  describing respectively the pairs of points that are similar to each other (eg., share the same label) and the pairs  which are dissimilar (eg., have different labels), Xing and al.~\cite{Xing2003} learn a matrix $M\succ 0$ by gradient descent such that the total pairwise distance in $\calD$ in maximized, while keeping the total pairwise distance in $\calS$ constant. 
While good performances are experimentally obtained, this MMC method tends to cluster similar points together and may thus perform poorly in the case of multi-modal data.
Furthermore, MMC requires two computationally costly projections at each gradient step: 
One projection on the cone of positive semi-definite matrices, and the other projection on the set of constraints.

 LMNN~\cite{WeinbergerSaul2006}  on the other hand is a projection-free metric learning method.
LMNN learns a global Mahalanobis metric using triplet information: For each point, we take as input a set of $k$ \textit{target neighbors} which should be brought close by the learned metric, while enforcing a unit margin with respect to points which are differently labelled. Contrary to MMC, LMNN handles well multi-modal data, but would optimally require oracle information of which points should be considered as targets of a given point. In practice, this is achieved by computing for each point the list of its $k$ nearest neighbors according to euclidean distance beforehand, but in specific applications the ``point neighborhoods'' can be gained using additional structural properties of the problems at hand.

While we consider in the remainder the LMNN framework, another more flexible approach in metric learning consists in learning {\em local} metrics, which allow to obtain a non-linear pseudo-metric while staying in a Mahalanobis framework\footnote{In Riemannian geometry, the distance is a geodesic length
$L(\gamma)=\int_a^b \sqrt{g_{\gamma(t)}(\dot\gamma(t),\dot\gamma(t))} \dt$ that can be interpreted as locally integrating Mahalanobis infinitesimal distances:
$L(\gamma)=\int_a^b D_{g(\gamma(t))}(\dot\gamma(t),\dot\gamma(t)) \dt$ for a metric tensor $g$.}, at the cost of greatly amplifying spatial complexity. Therefore, most works on the subject try to obtain a sparse encoding of such metrics.
For example, in~\cite{FetayaUllman2015}, Fetaya and Ullman learn one Mahalanobis metric per data point using only negative examples (eg., only information on dissimilarity), and obtain sparse metrics thanks to an equivalence with Support Vector Machines (SVMs). 
In \cite{ShiBelletSha2014}, Shi et. al. sparsely combine low-rank (one-dimensional) local metrics into a global metric. 
For a comprehensive survey on local metric learning, we refer the reader to~\cite{RamananBaker2011}.

\subsection{Large Margin Nearest Neighbors (LMNN)\label{sec:LMNN-M}}

Given a labeled input data-set $\calP=\{(x_1,y_1),\ldots,(x_1,y_1)\}$ of $n$ points $x_1,\ldots, x_n$ of $\bbR^d$, the 
Large Margin Nearest Neighbors\footnote{\url{http://www.cs.cornell.edu/~kilian/code/lmnn/lmnn.html}} (LMNN)~\cite{WeinbergerSaul2006} learns a Mahalanobis distance (ie., matrix $M\succ 0$). 
Since the $k$-NN classification does not change by taking any monotonically increasing function of the base distance (like its square), 
it is often more convenient mathematically to use the squared Mahalanobis distance that get rid of the square root.  
However, the squared Mahalanobis distance does not satisfy the triangle inequality. (It is a Bregman divergence~\cite{BD-2005,BVD-2007,BVD-2010}.)

In LMNN, for each point, we take as input the set of $k$ \textit{target neighbors} which should be brought close by the learned metric, while enforcing a unit margin with respect to points which have different labels.

To define the objective cost function~\cite{WeinbergerSaul2006} in LMNN, we consider two sets $\calS$ and $\calR$, or target neighbors and impostors:

\begin{itemize}
\item Distance of each point to its  \textit{target neighbors} shrink, $\epsilon_{\pull}(L)$:

\begin{equation}
\calS=\{(x_i,x_j) \st y_i=y_j \text{\ and\ } x_j\in N(x_j)\},
\end{equation}
where $N(x)$ denotes the neighbors of point $x$.

\item Keep a distance margin of each point to its \textit{impostors},  $\epsilon_{\push}(L)$:
\begin{equation}
\calR=\{(x_i,x_j,x_l) \st  (x_i,x_j)\in\calS \text{\ and\ }  y_i\not =y_l \}
\end{equation}

\end{itemize}

Using Cholesky decomposition $M= L^\top L\succ 0$, the LMNN cost function~\cite{WeinbergerSaul2006} is then defined as:
\begin{eqnarray}
\epsilon_{\pull}(L) &=& \Sigma_{i,i\rightarrow j} \Vert L(x_i-x_j) \Vert^2,\\
\epsilon_{\push}(L) &=& \Sigma_{i,i\rightarrow j} \Sigma_j (1-y_{il}) \left[ 1 + \Vert L(x_i-x_j) \Vert^2 - \Vert L(x_i-x_l) \Vert^2 \right]_+,\\
\epsilon(L) &=& (1-\mu)\epsilon_{\pull}(L) + \mu\epsilon_{\push}(L),
\end{eqnarray}
where $[x]_+=\max(0,x)$ and  $\mu$ is a trade-off parameter for tuning target/impostor relative importance, 
and $i\rightarrow j$ indicates that $x_j$ is a target neighbor of $x_i$. 
We define $y_{il} = 1$ if and only if $x_i$ and $x_j$ have same label,   $y_{il} = 0$ otherwise.

Thus the training of the Mahalanobis matrix $M = L^\top L$ is done by minimizing a linear combination of a \textit{pull} function which brings points closer to their target neighbors with a \textit{push} function that keeps the impostors away by penalizing the violation of the margin with a hinge loss.

The LMNN cost function is convex and piecewise linear~\cite{WeinbergerSaul2006}.
Replacing the hinge loss by slack variables, we obtain a semidefinite program, which allows us to solve the minimization problem with standard solver packages.

Instead, Weinberger and Saul~\cite{WeinbergerSaul2006} propose   a gradient descent where the set of \text{impostors} is re-computed every $10$ to $20$ iterations. 

In our implementation, we optimize the cost function by gradient descent: 
\begin{equation}
\epsilon(L_{t+1})=\epsilon(L_t)-\gamma\frac{\partial \epsilon(L_t)}{\partial L},
\end{equation}
where $\gamma>0$ is the {\em learning rate}, and:
\begin{equation}
\frac{\partial \epsilon}{\partial L} = (1-\mu)\Sigma_{i,i\rightarrow j} C_{ij} + \mu \Sigma_{(i,j,l) \in \calR_t} (C_{ij}-C_{il})
\end{equation}
with $C_{ij} = (x_i-x_j)^\top (x_i-x_j)$.

LMNN  is  a projection-free metric learning method that is quite easy to implement.
There is no projection mechanism like for the Mahalanobis Metric for Clustering (MMC)~\cite{Xing2003} method.

We shall now consider extensions of the LMNN method to Cayley-Klein elliptic~\cite{Bi2015} and hyperbolic geometries.


\subsection{Elliptic Cayley-Klein LMNN\label{sec:LMNN-E}}

Bi et al.~\cite{Bi2015} consider the extension of LMNN to the case of elliptic Cayley-Klein geometry.   
The cost function is defined as:
\begin{equation} 
\epsilon(L) = (1-\mu)\sum_{i,i\rightarrow j} d_E(x_i,x_j)+\mu \sum_{i,i\rightarrow j}\sum_l (1-y_{il})\zeta_{ijl}
\end{equation}
with  
\begin{equation}
\zeta_{ijl} = \left[1+d_E(x_i,x_j)-d_E(x_i,x_l)\right]_+.
\end{equation}

The gradient\footnote{There is  minor error in the expression of $\frac{\partial\epsilon(L)}{\partial L}$
in the original paper of Bi et al.~\cite{Bi2015}, as $C_{ij} + C_{ji}$ was replaced by $2C_{ij}$, which cannot be  the distance gradient that must be symmetric with respect to $x_i$ and $x_j$.}
 with respect to lower triangular matrix $L$ is computed as:
\begin{equation}\label{grad:ECK}
\frac{\partial\epsilon(L)}{\partial L} = (1-\mu)\sum_{i,i\rightarrow j}\frac{\partial d_E(x_i,x_j)}{\partial L}+\mu\sum_{i,i\rightarrow j}\sum_l (1-y_{il})\frac{\partial\zeta_{ijl}}{\partial L},
\end{equation}
with $C_{ij} = (x_i^\top,1)^\top (x_j^\top,1)$.

The gradient terms of Eq.~\ref{grad:ECK} are calculated as follows:
\begin{eqnarray}
\frac{\partial d_E(x_i,x_j)}{\partial L} &=& \frac{k}{\sqrt{S_{ii}S_{jj} - S_{ij}^2}}L\left(\frac{S_{ij}}{S_{ii}}C_{ii} + \frac{S_{ij}}{S_{jj}}C_{jj} - \left(C_{ij} + C_{ji} \right)\right)\\
\frac{\partial\zeta_{ijl}}{\partial L} &=& \begin{cases}
  \frac{\partial d_E(x_i,x_j)}{\partial L} -\frac{\partial d_E(x_i,x_l)}{\partial L} , & \text{if }  \zeta_{ijl} \geq 0, \\
  0, & \text{otherwise}.
\end{cases}
\end{eqnarray}

The elliptic LMNN loss is {\em not} convex, and thus the performance of the algorithm greatly depends on the  chosen initialization for $M = L^\top L$.
We may initialize the elliptic CK-LMNN either by  the sample mean  $m=\frac{1}{n}\sum_i x_i$ of the point set $\calP$,
and either the precision matrix  (inverse covariance matrix) of $\calP$ or the matrix obtained by Mahalanobis-LMNN.
We then build initial matrix $S$ as follows: 
 
\begin{equation}
G_{+} = 
\begin{pmatrix}
\Sigma & -\Sigma m\\
-m^\top \Sigma & m^\top \Sigma m + \frac{1}{\kappa^2}
\end{pmatrix}.
\end{equation}

Such a matrix is called a {\em generalized Mahalanobis matrix} in~\cite{Bi2015}. We term them curved Mahalanobis matrices.

Note that elliptic Cayley-Klein geometry are defined on the full domain $\mathbb{R}^d$, and furthermore the elliptic distance is bounded (by $\pi$ when $\kappa=1$).

\subsection{Hyperbolic Cayley-Klein LMNN\label{sec:LMNN-H}}

To ensure that the $(d+1)\times(d+1)$-dimensional matrix $S$ keeps the correct signature  $(1,0,d)$ during the LMNN gradient descent, we decompose $S=L^\top D L$ (with $L\succ 0$) and
perform a gradient descent on $L$ with the following gradient:

\begin{equation}
\frac{\partial d_H(x_i,x_j)}{\partial L} = \frac{k}{\sqrt{S_{ij}^2 - S_{ii}S_{jj} }}DL\left(\frac{S_{ij}}{S_{ii}}C_{ii}
 + \frac{S_{ij}}{S_{jj}}C_{jj} - \left(C_{ij} + C_{ji} \right)\right).
\end{equation}

We initialize $L=\begin{pmatrix}
L' & \\
&1\\
\end{pmatrix}$ and $D$ so that  $\calP\in\mathbb{D}_S$ as follows:
 Let  $\Sigma^{-1} = L'^\top L'$ (eg., by taking precision matrix $\Sigma^{-1}$ of $\calP$), and then choose the diagonal matrix as:

\begin{equation}
D = \begin{pmatrix}
-1 & \\
& \ddots\\
&&-1\\
&&& \kappa \ \max_x\Vert L'x \Vert^2
\end{pmatrix},
\end{equation}
with $\kappa>1$.

Let $\mathbb{D}_{S_t}$ denote the domain at a given iteration $t$ induced by the bilinear form $S_t$.
It may happen that the point set $\calP\not\in \mathbb{D}_{S_t}$ since we do not know the optimal learning rate $\gamma$ beforehand, and thus might have overshoot the domain.
When this case happens, we reduce $\gamma\leftarrow \frac{\gamma}{2}$, otherwise when the point set $\calP$ is fully contained inside the real conic domain, 
we let $\gamma\leftarrow 1.01\gamma$.

Like in the elliptic case, we  initialize the  hyperbolic CK-LMNN either by calculating the sample mean  $m=\frac{1}{n}\sum_i x_i$ of the point set $\calP$,
and either the precision matrix of $\calP$ or the matrix obtained by Mahalanobis-LMNN.
We then build initial matrix $S$ as follows: 
 
\begin{equation}
G_{-} = 
\begin{pmatrix}
\Sigma & -\Sigma m\\
-m^\top \Sigma & m^\top \Sigma m - \frac{1}{\kappa^2}
\end{pmatrix}.
\end{equation}

\begin{figure}%
\centering
\begin{tabular}{cc}
\includegraphics[width= 0.40\textwidth]{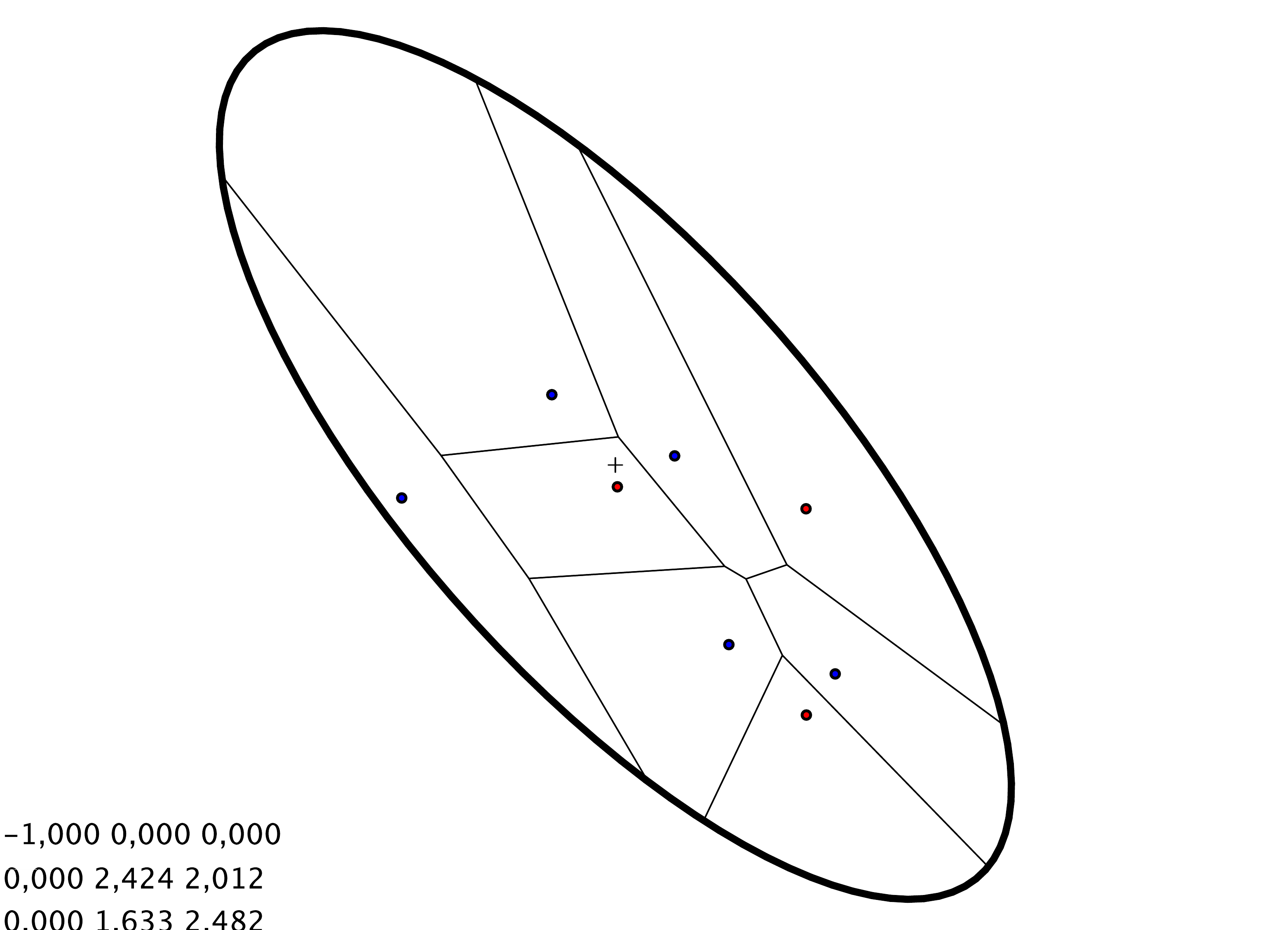} & \includegraphics[width= 0.40\textwidth]{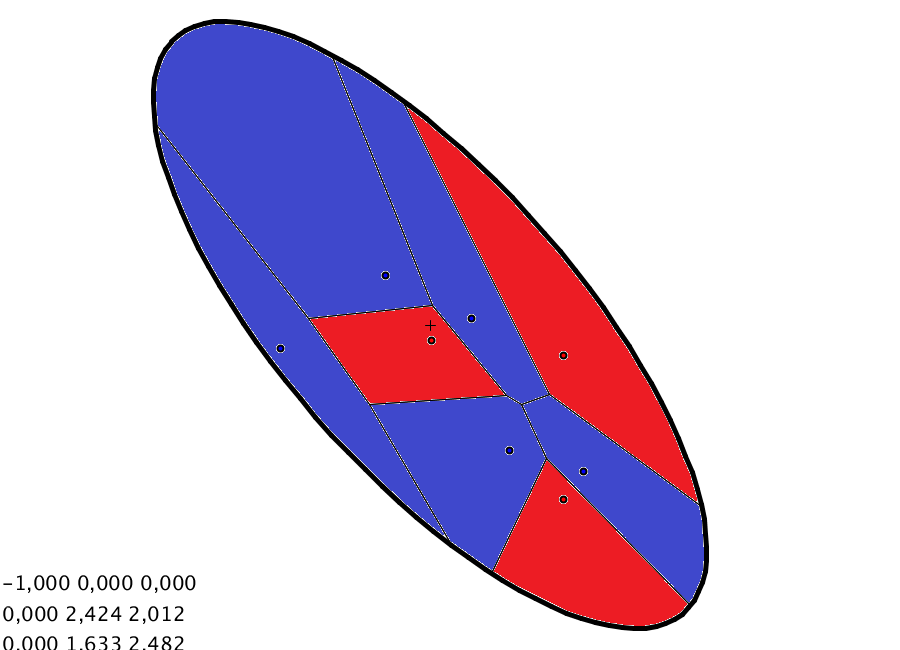}\\
(a) & (b)\\
\end{tabular}

\caption{Nearest neighbor ($k=1$) classification: (a) Binary labels of a point set shown in red/blue colors, and (b) bichromatic hyperbolic Cayley-Klein Voronoi diagram. The decision frontier is piecewise linear.}%
\label{fig:hckclassif}%
\end{figure}

Figure~\ref{fig:hckclassif} displays a hyperbolic Cayley-Klein Voronoi diagram for a set of $8$ generators (with labels $\pm 1$ displayed in blue/red), and
the bichromatic Voronoi diagram in case of binary classification. Notice that the {\em decision frontier} of  the  nearest-neighbor classifier ($k=1$) is the union of Voronoi facets (in 2D, edges) supporting different label cells. A similar result holds for the Cayley-Klein $k$-NN classifier: Its decision boundary is piecewise linear since the bisectors are (clipped) hyperplanes.

\section{Experimental results\label{sec:res}}

We report on our experimental results on some UCI data-sets.\footnote{\url{https://archive.ics.uci.edu/ml/datasets.html}}
Descriptions of those labelled data-sets are concisely summarized in Table~\ref{tab:uci}.

\begin{table}
	\begin{center}
	\caption{Characteristics of the UCI data-sets.}
	\begin{tabular}{|c|c|c|c|}
	\hline
	Data-set & \# Data points & \# Attributes & \# Classes\\
	\hline
	Wine & 178 & 13 & 3\\
	Sonar & 208 & 60 & 2\\
	Vowel & 528 & 10& 11\\
	Balance & 625 & 4 & 3\\
	Pima & 768 & 8 & 2\\
	\hline
	\end{tabular}
	\end{center}
	\caption{UCI data-sets chosen for the experiments.\label{tab:uci}}
\end{table}

We performed $k=3$ nearest neighbor classification.

As in \cite{Bi2015}, we performed leave-one-out cross validation for the {\tt wine} data-set, 
whereas for {\tt balance}, {\tt pima} and {\tt vowel} data-sets, we trained the model on random subsets of size $250$, testing it on the remaining data and repeating this procedure   $10$ times.

\begin{table}[ht]
\begin{center}

  \begin{tabular}{|c|c|c|c|c|}
  \hline
  k & Data-set & elliptic & Hyperbolic & Mahalanobis\\
  \hline
  1& wine & \textbf{0.989}& 0.865 & 0.984\\
  & vowel & \textbf{0.832}& 0.797 & 0.827\\
  & balance & \textbf{0.924}& 0.891 & 0.846\\
  & pima &\textbf{0.726} &0.706 & 0.709\\
  \hline
  3& wine & 0.983 & 0.871& \textbf{0.984}\\
  & vowel & \textbf{0.828} & 0.782& 0.827\\
  & balance & \textbf{0.917} & 0.911&0.846\\
  & pima &0.706 &0.695& \textbf{0.709}\\
  \hline
  5& wine & 0.983& & \textbf{0.984}\\
  & vowel & 0.826& 0.805& \textbf{0.827}\\
  & balance & \textbf{0.907}& 0.895& 0.846\\
  & pima &\textbf{0.714} &0.712& 0.709\\
  \hline
   11& wine & \textbf{0.994} & 0.983& 0.984\\
  & vowel & \textbf{0.839} & 0.767& 0.827\\
  & balance & 0.874 &\textbf{0.897} &0.846\\
   & pima &\textbf{0.713} &0.698& 0.709\\
  \hline 
  \end{tabular}
  \end{center}
	\caption{Experiment results for $3$-NN LMNN classification.\label{tab:exp}}
\end{table}

We observe that the elliptic CK-LMNN performs quite  better than the Mahalanobis LMNN and the hyperbolic CK-LMNN.

\subsection{Spectral decomposition and  proximity queries in Cayley-Klein geometry\label{sec:proximity}}

To avoid to compute $d_E$ or $d_H$ for arbitrary matrix $S$, we apply the {\em matrix factorization} (elliptic case $S = L^\top L$, or hyperbolic case   $S=L^\top D L$ ) and perform coordinate changes so that it is enough to consider the canonical metric distances:
\begin{eqnarray}
d_E(x',y') &=& \arccos\left(\frac{\inner{x'}{y'}}{\Vert x'\Vert\Vert y'\Vert}\right),\\
d_H(x',y') &=&\arccosh \left(\frac{1-\inner{x'}{y'}}{\sqrt{1-\inner{x'}{x'}}\sqrt{1-\inner{y'}{y'}}}\right).
\end{eqnarray}

Alternatively, consider the {\em spectral decomposition} of matrix $S=O \Lambda O^\top$ obtained by eigenvalue decomposition (with diagonal matrix $\Lambda=\diag(\Lambda_{1,1},\ldots,\Lambda_{d+1,d+1})$), and let us write canonically:
\begin{equation}
S = O D^{\frac{1}{2}} {\left[\begin{array}{cc} I & 0 \cr 0 & \lambda  \end{array}\right]} D^{\frac{1}{2}}  O^\top,
\end{equation}
where $\lambda=\in\{-1,1\}$ and $O$ is an orthogonal matrix (with $O^{-1}=O^\top$).
 The diagonal matrix $D$ has {\em all} positive values, with $D_{i,i}=\Lambda_{i,i}$ and $D_{d+1,d+1}=|\Lambda_{d+1,d+1}|$ so that $ D^{\frac{1}{2}}$ is defined
as the diagonal matrix obtained by taking  element-wise the square root values of the matrix.

We rewrite the bilinear form into a canonical  form by mapping the points
$x$ to 
$\tilde{x'}=D^{\frac{1}{2}} O^\top \begin{pmatrix}x \\ 1\end{pmatrix} =\begin{pmatrix}x'' \\ w\end{pmatrix}$. 
Since $\tilde{x'} =  \begin{pmatrix}x' \\ 1\end{pmatrix}$, we can then find $x'=\frac{x''}{w}$.
When $\lambda>0$ (elliptical case with $D_{d+1,d+1}>0$), we have $S_S(p,q)=S_E(p',q')=S_I(p',q')$.
When $\lambda<0$ (hyperbolic case with $D_{d+1,d+1}<0$), we have $S_S(p,q)=S_H(p',q')$, with $H=\diag(1, ..., 1, -1)$ the canonical matrix form for hyperbolic Cayley-Klein spaces.

Notice that in the ordinary Mahalanobis case, instead of using the Cholesky decomposition, we may also use the $L_1 D {L}_1^\top$ matrix decomposition where
$L_1$ is a unit lower triangular matrix (with diagonal elements all $1$), and $D$ is a diagonal matrix of positive elements.
The mapping is then $x'=D^{\frac{1}{2}} L_1^\top$ or $x'= (L_1 D^{\frac{1}{2}})^\top$ since $D=D^\top$.
Thus by transforming the input space into one of the {\em canonical} Euclidean/elliptical/hyperbolic spaces, we avoid to perform costly matrix multiplications required 
in the general bilinear form, and once the structure (say, a $k$-NN decision boundary or a Voronoi diagram) has been recovered, we can map back to the original space (say, for classifying new observations using the original coordinate system).

Nearest neighbor proximity queries can then be answered using various spatial data-structures.
For example, we may consider the Vantage Point Tree data-structures~\cite{VPT-1993,VPT-2009}.

\begin{figure}%
\centering
\includegraphics[scale = 0.6]{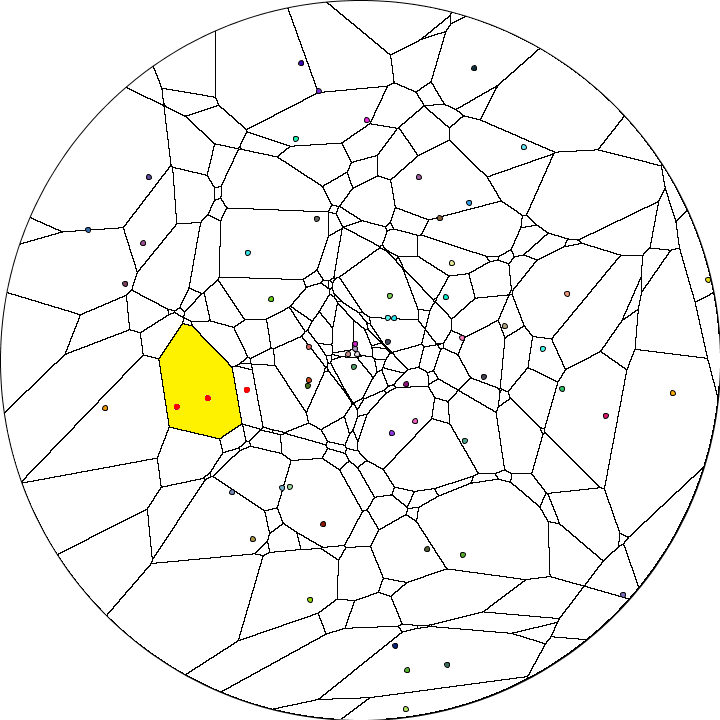} 
\caption{Example of a $3$-order affine hyperbolic Voronoi diagram. For each point in the  yellow cell the common three closest nearest neighbors are displayed as red points.}%
\label{fig:CKVorK3}%
\end{figure}

In small dimensions, we can compute the $k$-order elliptic/hyperbolic Voronoi affine diagram, as depicted in Figure~\ref{fig:CKVorK3}.
The $k$-order Voronoi diagram is affine since the bisectors are affine.
Neighbor queries can then be reported efficiently in logarithmic time in 2D after preprocessing time, see~\cite{CG-2000} for further details.

\subsection{Mixed curved Mahalanobis distance\label{sec:mixedres}}

We consider the {\em mixed} elliptic/hyperbolic Cayley-Klein distance:
\begin{equation}
d(x,y) = \alpha d_E(x,y) + (1-\alpha) d_H(x,y).
\end{equation}

Since the sum of (Riemannian) metric distances is a (Riemannian) metric distance, we deduce that $d(x,y)$ is a (Riemannian) metric distance.
However, this ``blending'' of positive with negative constant curvature (Riemannian) geometries does not yield a constant curvature (Riemannian) geometry.
Indeed, although that  the metric tensors blend locally, the {\em Ordinary Differential Equation} (ODE) characterizing the geodesics solves differently.

Notice that we mix a bounded distance (elliptic CK) with an unbounded distance (hyperbolic CK) via the hyperparameter $\alpha$ that needs to be tuned.
Table~\ref{res:mix} shows the preliminary experimental results. Those results indicate better performance for the mixed model in most (but not all) cases.
This should not be surprising as  a smooth non-constant Riemannian manifold will better model data-sets than a constant-curvature manifold.

 	\begin{table}
\begin{center}
  \label{tab:mixed}
  \begin{tabular}{|c|c|c|c|c||c|c|}
  \hline
    Datasets & Mahalanobis & elliptic & Hyperbolic & Mixed& $\alpha$ & $\beta=(1-\alpha)$ \\
    \hline
    Wine & \textbf{0.993} & 0.984 & 0.893 & 0.986 & 0.741 & 0.259\\
    Sonar & 0.733 & 0.788 & 0.640 & \textbf{0.802} & 0.794 & 0.206\\
    Balance & 0.846	& 0.910 &	0.904 &	\textbf{0.920} & 0.440 & 0.560\\
    Pima & 0.709  &	0.712 &	0.699&	\textbf{0.720} & 0.584 & 0.416\\
    Vowel & 0.827 &	0.825 &	0.816 &	\textbf{0.841} & 0.407 & 0.593\\
    \hline
	\end{tabular}
	\end{center}
	\caption{Experimental classification results on mixed curved Mahalanobis distances.\label{res:mix}}
	\end{table}

\section{Conclusion and perspectives\label{sec:concl}}

We  considered Cayley-Klein geometries for super-vised classification purposes in machine learning.
First, we   studied some  nice properties of the Voronoi diagrams and balls in Cayley-Klein geometries:
We proved that the  Cayley-Klein Voronoi diagram is affine, and reported formula to build it as an equivalent (clipped) power diagram.
We then showed that  Cayley-Klein  balls have Mahalanobis shapes with displaced centers, and gave the explicit conversion formula.
Second, we extended the LMNN  framework to hyperbolic Cayley-Klein geometries that were not considered in~\cite{Bi2015}, and proposed learning a mixed elliptic/hyperbolic distance that experimentally shows good improvement over constant-curvature Cayley-Klein geometries.

The fact that the  Cayley-Klein  bisectors are hyperplanes offers nice computational perspectives in machine learning and computational geometry.
For example, it would be interesting to study Multi-Dimensional Scaling~\cite{MDS-CK-1979} or Support Vector Machines (SVMs) in Cayley-Klein geometries, or to mesh anisotropically~\cite{anisotropicMeshing-2015} in Cayley-Klein geometries.

\vskip 0.5cm
Supplemental information is available online at:\\
\centerline{\url{https://www.lix.polytechnique.fr/~nielsen/CayleyKlein/}}


\end{document}